\documentclass{article} % For LaTeX2e
\usepackage{iclr2025_conference,times}

\usepackage{amsmath,amsfonts,bm}

\def\eqref#1{equation~\ref{#1}}
\def\1{\bm{1}}

\DeclareMathAlphabet{\mathsfit}{\encodingdefault}{\sfdefault}{m}{sl}
\SetMathAlphabet{\mathsfit}{bold}{\encodingdefault}{\sfdefault}{bx}{n}

\usepackage{hyperref}
\usepackage{url}
\usepackage{graphicx}
\usepackage{xspace}
\usepackage{booktabs}
\usepackage{multirow}
\usepackage{paralist}

\newcommand{\sysname}{{FlexCAD}\xspace}
\newcommand\fref[1]{Fig.~\ref{#1}}

\newcommand{\eg}{\emph{e.g.},\xspace}
\newcommand{\ie}{\emph{i.e.},\xspace}
\newcommand{\etc}{etc.\xspace}

\newcommand\tabref[1]{Table~\ref{#1}}

\definecolor{cloop}{rgb}{0,176/255,80/255}
\definecolor{cf1}{rgb}{0.27,0.596,0.8}
\definecolor{cf4}{rgb}{0.184,0.4,0.678}
\definecolor{b1}{rgb}{0.929,0.373,0.267}
\definecolor{b2}{rgb}{0.208,0.141,0.996}
\title{\sysname: Unified and Versatile Controllable CAD Generation with Fine-tuned Large Language Models}

\author{
Zhanwei Zhang$^{1}$, ~Shizhao Sun$^{2,\dag}$, ~Wenxiao Wang$^{3,\dag}$, ~Deng Cai$^{1}$, ~Jiang Bian$^{2}$
\\[.253cm]
{
\textsuperscript{1} {State Key Lab of CAD\&CG, Zhejiang University} \;
\textsuperscript{2} {Microsoft} \;
\textsuperscript{3} {School of Software Technology, Zhejiang University} \;
}
}

\author{{Zhanwei Zhang}$^{1}$\thanks{{ }Work done during an internship at Microsoft Research Asia.}\ , 
~Shizhao Sun$^{2}$\thanks{{ }Corresponding author.}\ , ~Wenxiao Wang$^{3\dag}$, ~Deng Cai$^{1}$, ~Jiang Bian$^{2}$
\\
$^{1}$ State Key Lab of CAD\&CG, Zhejiang University \\ 
$^{2}$ Microsoft Research \\
$^{3}$ School of Software Technology, Zhejiang University \\
\texttt{\{zhanweizhang, wenxiaowang\}@zju.edu.cn } \\
\texttt{\{shizsu, jiabia\}@microsoft.com, {dengcai@cad.zju.edu.cn}
}}

\iclrfinalcopy % Uncomment for camera-ready version, but NOT for submission.
\begin{document}

\maketitle

\begin{abstract}
Recently, there is a growing interest in creating computer-aided design (CAD) models based on user intent, known as controllable CAD generation.
Existing work offers limited controllability and needs separate models for different types of control, reducing efficiency and practicality.
To achieve controllable generation across all CAD construction hierarchies, such as sketch-extrusion, extrusion, sketch, face, loop and curve,
we propose \sysname, a unified model by fine-tuning large language models (LLMs).
First, to enhance comprehension by LLMs, we represent a CAD model as a structured text by abstracting each hierarchy as a sequence of text tokens.
Second, to address various controllable generation tasks in a unified model, we introduce a hierarchy-aware masking strategy.
Specifically, during training, we mask a hierarchy-aware field in the CAD text with a mask token.
This field, composed of a sequence of tokens, can be set flexibly to represent various hierarchies.
Subsequently, we ask LLMs to predict this masked field.
During inference, the user intent is converted into a CAD text with a mask token replacing the part the user wants to modify, which is then fed into \sysname to generate new CAD models. 
Comprehensive experiments on public dataset demonstrate the effectiveness of \sysname in both generation quality and controllability.
Code will be available at \url{https://github.com/microsoft/FlexCAD}.
\end{abstract}

\begin{figure*}[h]
\centering
\includegraphics[width=\textwidth]{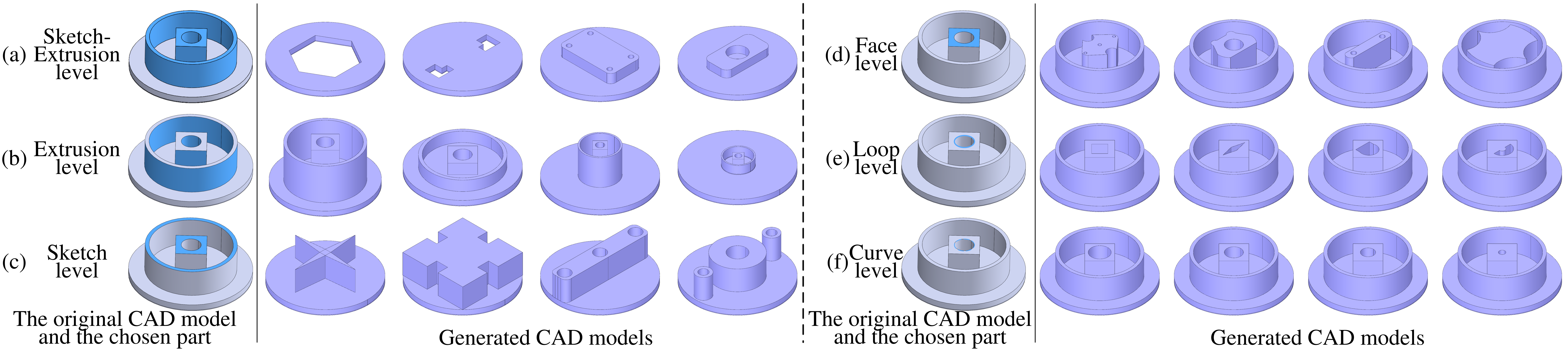}
\caption{Controllable CAD generation achieved by \sysname. In each sub-figure, the left side shows the input: an original CAD model along with the part the user intends to modify (highlighted in \textcolor{cf1}{blue}). The right side displays the output: multiple new CAD models with only the chosen part changed. Users have the flexibility to specify the part in any CAD construction hierarchies, ranging from coarse levels like sketch-extrusion to fine levels like curve (as illustrated from (a) to (f)).}
\label{fig1}
\end{figure*}
\section{Introduction}
A computer-aided design (CAD) model is a digital representation of a 2D or 3D object.
It has been widely used across numerous industries, including architecture, product design and manufacturing, facilitating precise, efficient, and innovative development~\cite{ganin2021computer,khan2024cad}.
In commonly used CAD tools like SolidWorks and AutoCAD, \emph{sketch-and-extrude modeling} (SEM) is prevalent.
This involves drawing 2D sketches and then extruding them into 3D shapes. 
Compared to other representations, such as Constructive Solid Geometry (CSG)~\cite{yu2024d}, B-rep~\cite{xu2024brepgen}, 
or voxel~\cite{li2023secad} and point cloud~\cite{khan2024cad}-based formats, SEM, incorporating multiple CAD construction hierarchies including \emph{sketch-extrusion}, \emph{extrusion}, \emph{sketch}, \emph{face}, \emph{loop} and \emph{curve} (see \fref{fig3}(a)),
directly illustrates the drawing process of a 3D object. 
This allows for easy editing and reuse of CAD models, which is essential in CAD development.

Recently, there is an increasing interest in developing generative models to automatically produce SEM of a CAD model\footnote{In the following, we will use CAD model to refer to SEM of a CAD model for brevity.}.
Specifically, DeepCAD~\cite{wu2021deepcad} focuses on uncontrollable generation, where a CAD model is generated from a randomly sampled vector.
However, providing controllability, \ie generating CAD models according to user intent, is crucial for the practical application of generative models. 
To address this, SkexGen~\cite{pmlr-v162-xu22k} and Hnc-cad~\cite{pmlr-v202-xu23f} implement disentangled codebooks to offer some levels of control.
As each codebook encodes a particular construction hierarchy, their controllability is quite restricted.
For instance, SkexGen does not allow selecting a specific sketch for modifications when a CAD model comprises multiple sketches, nor can it handle finer-grained hierarchies such as faces and loops.
Hnc-cad lacks control over the topology and geometry of curves.
In summary, existing methods face challenges in providing adequate controllability across all CAD construction hierarchies.
Additionally, they require separate models to deliver different types of control, which is inefficient and less practical.

The emergence of large language models (LLMs) offers insights for addressing these challenges.
First, LLMs have exhibited remarkable success in handling diverse user queries with a single and unified model~\cite{chung2024scaling}.
This phenomenon not only occurs in natural language tasks but also extends to other areas with domain-specific fine-tuning, such as human motion generation~\cite{jiang2024motiongpt} and crystal material synthesis~\cite{gruver2024finetuned}.
Second, LLMs might have acquired CAD-related knowledge during the pre-training by learning CAD-specific codes, such as JSCAD codes~\cite{makatura2023can}.
Third, prior to the rise of LLMs, small transformer-based models were explored for tasks like uncontrollable generation and image-to-sketch translation in the 2D sketch domain~\cite{ganin2021computer}, showcasing the possibility of LLMs from a different perspective.

In this work, we introduce \sysname, a unified model designed for controllable CAD generation across all hierarchies by fine-tuning LLMs.
As shown in \fref{fig1}, \sysname receives the original CAD model along with the part the user wants to modify (highlighted in blue). 
Here, users can specify the part in any hierarchy.
\sysname then generates multiple new CAD models, altering only the selected part.
To achieve these abilities, first, \sysname translates a CAD model into a \emph{concise and structured text} (see \fref{fig3}). 
Specifically, in each sketch, the curve type (\eg a line) is directly represented as textual tokens. 
The numerical data indicating geometry (\eg point coordinates in a line) is converted into decimal integers and then into textual tokens. 
A special token is added to mark the end of each hierarchy.
Tokens from the finer-level hierarchy are concatenated to form the representation for the coarser-level hierarchy.
We use a similar way to convert each extrusion.
Consequently, unlike the one-hot representation used in~\cite{pmlr-v162-xu22k}, \sysname provides a concise text representation of a CAD model, facilitating easier processing and understanding by LLMs.
Second, \sysname introduces a \emph{hierarchy-aware masking} strategy to enable fine-tuning LLMs for various controllable CAD generation tasks (see \fref{fig2}).
During training, we replace a hierarchy-aware field, which contains a sequence of tokens in the CAD text, with a mask token.
This field can be set adaptably to reflect various hierarchies.
Then, we ask LLMs to predict the masked field.
To achieve this, we design prompt templates for all hierarchies, where the mask tokens are tailored to match the corresponding hierarchies.
These templates are uniformly sampled at each epoch during the fine-tuning of LLMs.
In this way, we ensure that the generation tasks for all hierarchies are learned in a single and unified model.
Besides, unlike~\cite{pmlr-v162-xu22k, pmlr-v202-xu23f} that requires multi-stage training, \sysname achieves end-to-end training.
During inference, a CAD model is represented as a CAD text with a mask token replacing the part the user wants to change.
The masked CAD text is fed into the fine-tuned LLMs to get predictions.
After infilling the masked text with these predictions,
\sysname produces CAD texts that can be rendered into new CAD models.
Overall, our contributions are:

\begin{compactitem}
\item 
We propose \sysname, a unified and versatile model for controllable CAD generation across all hierarchies, including sketch-extrusion, extrusion, sketch, face, loop and curve.
\item 
To the best of our knowledge, \sysname is the first to leverage LLMs for controllable CAD generation. It converts a CAD model into a brief, structured text and employs hierarchy-aware masking to fine-tune LLMs for various controllable CAD generation tasks.
\item 
We conduct extensive experiments on public datasets. 
Despite its simplicity, \sysname greatly improves generation quality and controllability, showing its effectiveness on the tasks presented in this work and indicating potential for other CAD generation scenarios. 
\end{compactitem}

\section{Related Work}
\label{gen_inst}
{\setlength{\parindent}{0cm}\textbf{CAD Model Generation.}}
A CAD model represents a 2D or 3D object digitally, with far-reaching applications spanning multiple industries~\cite{li2024sfmcad}.
Existing CAD generation methods can be classified into three categories based on representations of CAD models~\cite{khan2024cad}: constructive solid geometry (CSG), boundary representation (B-rep) and sketch-and-extrude modeling (SEM).
CSG combines primitives (\eg cubes, cylinders, or spheres) via Boolean operations (\eg union, subtraction or difference) to construct a CSG tree~\cite{laidlaw1986constructive,du2018inversecsg,kania2020ucsg,ren2021csg,yu2022capri,yu2024d}. 
B-rep characterizes a CAD model as a graph~\cite{ansaldi1985geometric}, consisting of sets of interconnected faces, edges, and vertices~\cite{jayaraman2023solidgen,wang2022neural,jayaraman2023solidgen,xu2024brepgen}.

Our work focuses on SEM, which directly models the drawing process of a CAD model, \ie drawing 2D curves to make sketches and then extruding them into 3D shapes.
There are multiple construction hierarchies in SEM, such as sketch-extrusion, extrusion, sketch, face, loop and curve.
Compared to CSG and B-rep, it enables easy editing and reuse of CAD models, which is critical in the CAD design process.
For SEM of a CAD model, DeepCAD~\cite{wu2021deepcad} utilizes a transformer-based~\cite{transformer} autoencoder for unconditional CAD generation.
It merely designates controllable generation as a future application.
Nevertheless, offering controllability, \ie the ability to generate CAD models based on user intents, is critical for enhancing the efficiency of CAD design workflow in practical applications.
SkexGen \cite{pmlr-v162-xu22k} proposes to disentangle the topology and geometry of sketches and extrusions to provide some levels of control.
However, it does not allow choosing a specific sketch, face, loop or extrusion for editing when multiple options exist.
Hnc-cad \cite{pmlr-v202-xu23f} exploits three codebooks based on VQ-VAE framework \cite{van2017neural} to control loop, sketch and extrusion level generation, respectively.
While it offers finer control, it struggles to manage the topology and geometry in the curve level.
To sum up, despite significant progress, existing work face challenges in offering controllability across all construction hierarchies.
Besides, they develop separate models to support different types of control.

{\setlength{\parindent}{0cm}\textbf{Large Language Models (LLMs).}} 
LLMs have shown significant success recently~\cite{touvron2023llama,liu2024visual}. 
Fine-tuning LLMs has demonstrated notable versatility and efficacy~\cite{luo2024cheap,chung2024scaling,roziere2023code,maftl,zhang2024llamaadapter,li2024finetuning}. 
Specifically, the fine-tuned LLMs have displayed exceptional success by simultaneously handling various generation tasks within a unified framework~\cite{wu2024unigen,jiang2024motiongpt,gruver2024finetuned,Zou_2024_CVPR}.
Moreover, LLMs are generally pre-trained using extensive high-quality datasets with cross-disciplinary and abundant knowledge~\cite{wenzek2020ccnet,touvron2023llama}. 
For example, LLMs may have acquired CAD-related knowledge by learning CAD codes, such as JSCAD codes~\cite{makatura2023can}.
Encouraged by these achievements, we explore the potential of using LLMs to tackle the aforementioned challenges in CAD generation.

\section{Methodology}
\begin{figure*}[h]
\centering
\includegraphics[width=\textwidth]{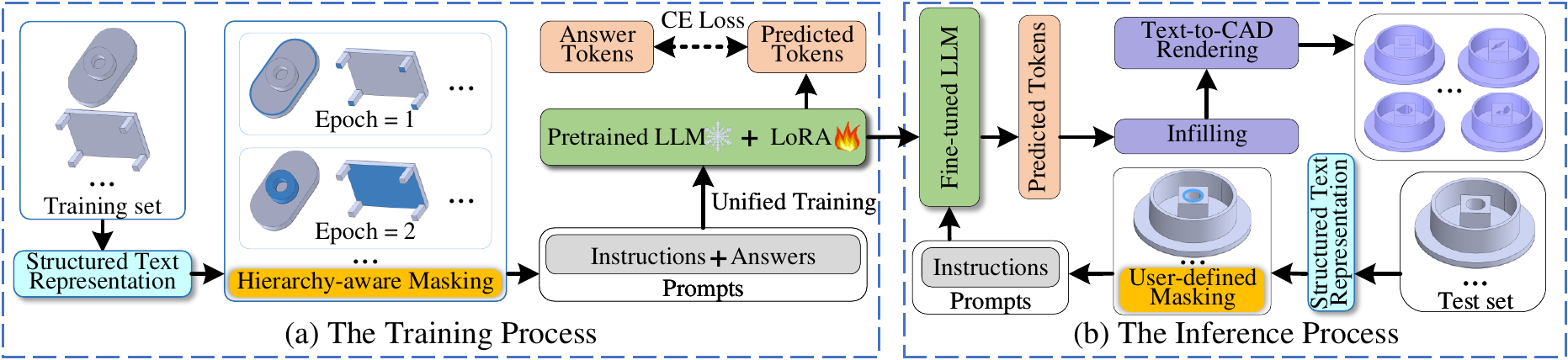}
\caption{The overall framework of \sysname.
(a) Training process.
Initially, a CAD model is converted into a structured text.
Next, a hierarchy-aware masking strategy is proposed to mask a specific field in the text with a special mask token.
This field is set differently at each epoch to reflect various hierarchies.
Then, LLMs are fine-tuned to predict the masked field.
(b) Inference process.
The original CAD model is transformed into a structured text with a mask token replacing the part the user wants to change. 
The fine-tuned LLMs are provided with this masked text to generate diverse predictions, which are then converted into new CAD models by infilling and rendering.}
\label{fig2}
\end{figure*}
In this section, we introduce \sysname, a unified model for controllable CAD generation across all construction hierarchies.
As shown in \fref{fig1}, it receives an original CAD model along with the part the user wants to modify (highlighted in blue), and generates multiple new CAD models with only the selected part altered.
To achieve this, as illustrated in \fref{fig2}, \sysname first translates a CAD model into a structured text (Sec.~\ref{3.1}), and then introduces a hierarchy-aware masking strategy to enable fine-tuning LLMs for multiple controllable CAD generation tasks (Sec.~\ref{3.2})

\subsection{Representing CAD as Structured Text}\label{3.1}
\begin{figure*}[!t]
\centering
\includegraphics[width=\textwidth]{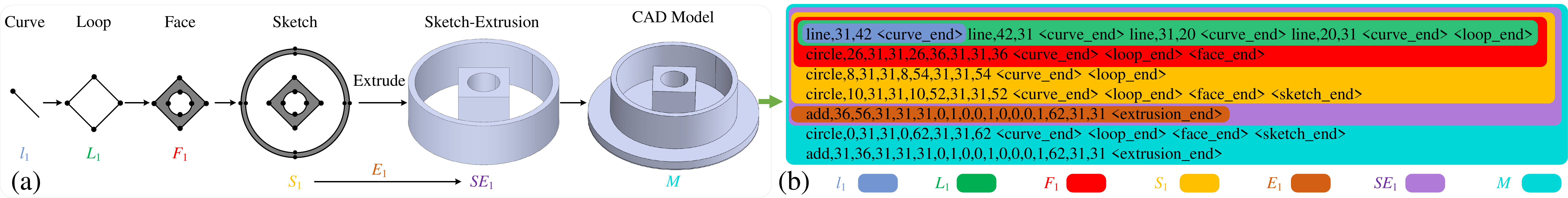}
\caption{(a) An illustration for construction hierarchies of a CAD model.
(b) Structured text representation for the CAD model shown in (a). The colors beneath the texts in (b) are used to indicate the relationship to construction hierarchies depicted in (a), \eg blue for a curve and green for a loop.
}
\label{fig3}
\end{figure*}
Following the conventional definition of SEM of a CAD model~\cite{wu2021deepcad, pmlr-v162-xu22k}, there are multiple construction hierarchies, as illustrated in \fref{fig3}(a).
% As shown in \fref{fig3}(a),
A `\emph{curve}', \ie line $l$, arc $a$, or circle $c$, forms the base level, represented by one, two, or four points respectively.
Each point is denoted by its $x$ and $y$ coordinates.
A `\emph{loop}' $L$ denotes a closed path, comprising either a single curve (\ie circle) or multiple curves (\eg line-arc-line).
A `\emph{face}' $F$ is a 2D area, characterized by a single loop or an outer loop with one or multiple inner loops acting as holes.
A `\emph{sketch}' $S$ is composed of one or multiple faces, sharing a common extrusion command.
An `\emph{extrusion}' $E$ is a command that extends a sketch from a 2D plane into a 3D body.
A `\emph{sketch-extrusion}' $SE$ represents a single sketch-extrusion 3D body. 
A `\emph{CAD model}' $M$ comprises one or multiple $SE$ entities.
As shown in \fref{fig3}(b), we represent a CAD model as a succinct and structured text.
Specifically, in each sketch, we start by representing a curve since it is the base level.
The curve type (\ie line, arc or circle) is represented directly as textual tokens.
The point coordinates of the curve, which are numerical, are expressed as decimal integers and then converted into textual tokens.
This contrasts with ~\cite{pmlr-v162-xu22k} that uses binary representation for point coordinates.
For example, when discretizing coordinates into a 64$\times$64 grid, \cite{pmlr-v162-xu22k} denotes the center coordinate as $([0, 1, 1, 1, 1, 1], [0, 1, 1, 1, 1, 1])$, while we represent it as $(31, 31)$.
Next, the curve is denoted as a sequence of textual tokens, with the first one indicating its type and the others representing point coordinates (the text with a blue background in \fref{fig3}(b)).
Notably, we add a special textual token `$H$\_end' to mark the end of each hierarchy, where $H\in$ \{curve, loop, face, sketch, extrusion\}.
This is also different from ~\cite{pmlr-v162-xu22k} where one-hot vectors are used as ending flags.
We concatenate tokens of multiple curves to create the representation for a loop (the text with a green background in \fref{fig3}(b)).
Then, we use a similar way to form representations of other hierarchies, including face and sketch (the texts with red and yellow backgrounds in \fref{fig3}(b)).
Furthermore, an extrusion can also be represented using textual tokens, with the first one specifying its type (\eg add or cut) and the others denoting its numerical attributes (the text with a brown background in \fref{fig3}(b)).
Finally, a complete CAD model is assembled by concatenating all the textual tokens from its sketch-extrusions (the text with a cyan background in \fref{fig3}(b), please see the meaning of numbers in the text in the appendix).
Consequently, we convert a CAD model into a structured text, enabling efficient processing and comprehension by LLMs.
Besides, our \sysname shortens the overall token length compared to~\cite{pmlr-v162-xu22k}.
Moreover, this text is straightforward to parse and interpret, thereby facilitating the implementation of the below hierarchy-aware masking strategy.

\subsection{Fine-tuning LLMs with Hierarchy-Aware Mask Prediction}\label{3.2}
\begin{figure*}[h]
\centering
\includegraphics[width=\textwidth]{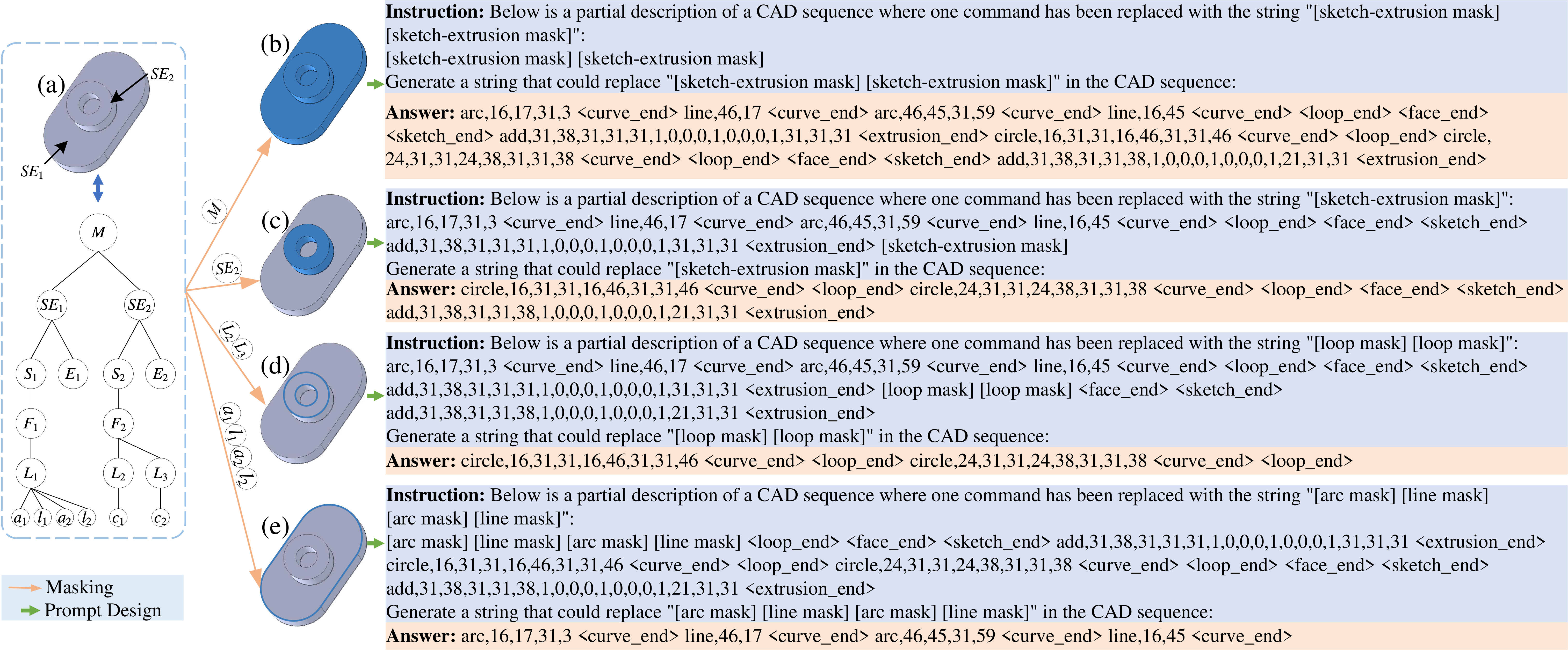}
\caption{
(a) illustrates a CAD model and its structural diagram. (b), (c), (d) and (e) are four examples for prompt templates with the mask tokens designed to represent different construction hierarchies. The masked field for different hierarchies in the CAD model are highlighted in \textcolor{cf4}{blue}.
}
\label{fig4}
\end{figure*}
In the following, with the structured text representation (referred to as CAD text for simplicity), we introduce how to fine-tune LLMs to develop a unified model for various controllable CAD generation tasks.
In general, during training, a hierarchy-aware field in the CAD text is replaced with a mask token.
The field, which consists of a sequence of tokens, can be designed to reflect different CAD construction hierarchies.
Next, LLMs are asked to predict the masked field (see \fref{fig2}(a)).
To accomplish this, we design different prompt templates, where the mask tokens are designed to align with the corresponding hierarchies (see \fref{fig4}).
During inference, given a CAD model, with a defined mask token, users can specify the part they want to modify (see \fref{fig2}(b)).
Below, we further detail the design of prompt templates, the unified training and inference processes.

{\setlength{\parindent}{0cm}\textbf{Prompt Template Design with Hierarchy-aware Mask.}} 
A prompt template includes an instruction with a special mask token replacing a hierarchy-aware field, and an answer containing the tokens of this field.
Specifically,
for the 
\textbf{CAD} level, we mask each internal sketch-extrusion with $\mathrm{[sketch\text{-}extrusion \ mask]}$.
In this case, other than the sketch-extrusion number, no information from the original CAD model is preserved.
\fref{fig4}(b) shows an example.
This allows us to freely generate CAD models with the expected number of sketch-extrusions during inference, facilitating the creation of CAD models with varying complexity.
For the \textbf{sketch-extrusion}, \textbf{sketch} and \textbf{extrusion} levels,
we replace the relevant field with $\mathrm{[sketch\text{-}extrusion \ mask]}$, $\mathrm{[sketch \ mask]}$, or $\mathrm{[extrusion \ mask]}$, respectively. \fref{fig4}(c) shows an example with one masked sketch-extrusion.
For the \textbf{face} (\textbf{loop}) levels, given a face (loop),
if it exclusively forms a sketch (face), 
we mask this face (loop) with $\mathrm{[face \ mask]}$ ($\mathrm{[loop \ mask]}$).
In cases where multiple faces (loops) belong to the same sketch (face), we use a corresponding number of mask tokens to mask them all at once.
\fref{fig4}(d) illustrates an example, where two loops are replaced by two mask tokens.
With this strategy, the model learns to generate faces (loops) with varying numbers as described in different instructions. 
For the \textbf{curve} level, all curves of the same loop are masked with their type indicated in the mask token (\ie line, arc or circle). 
\fref{fig4}(e) presents an example where four curves (arc-line-arc-line) belonging to the same loop are masked simultaneously.
As the curve is the fundamental hierarchical level, the control of the topology and geometry of a sketch comes from it.
Specifically, once trained and given a loop, by keeping its internal curve type and number unchanged, we can only modify the geometry. 
Alternatively, by varying the type or number of curves, we can alter the topology.

{\setlength{\parindent}{0cm}\textbf{Unified Training by Sampling Prompt Templates.}}\label{3.3}
At each epoch, for a given CAD text, we uniformly sample a prompt template from the above seven hierarchies. 
The instruction in the template asks LLMs to predict the masked field autoregressively. 
Then, the cross-entropy (CE) loss between the prediction and the answer in the template is back-propagated to update the LLMs.
To sum up, the advantage here is two-fold.
First, by randomly choosing existing prompt templates at each epoch, we aim to establish a unified controllable generation model for various hierarchies.
Second, beyond the existing prompt templates, we can incorporate new templates that support other tasks, such as unconditional generation.
Notably, we fine-tune LLMs using LoRA~\cite{hu2022lora} which allows a few parameters to be trainable while keeping most parameters fixed.
This allows us to leverage the advantages of large-scale models while accelerating model convergence \cite{hu2022lora}.

{\setlength{\parindent}{0cm}\textbf{Inference with User-defined Mask.}}
During inference, a CAD model is first converted to a CAD text, with a mask token replacing the part that needs modification.
This masked CAD text is then input into the fine-tuned LLMs to produce predictions.  
After infilling the masked text with these predictions, \sysname can provide various CAD texts that can be rendered into diverse CAD models.
Notably, users do not have to strictly adhere to the masking pattern defined in the training process.
For example, although the prompt template, used in training, masks all the loops tied to a face simultaneously, this is not mandatory in the inference.
Due to the strong generalization capability of LLMs, it is possible to only mask a single loop for local editing, as illustrated in \fref{fig2}(b).

\section{Experiments}
\label{exp}
\subsection{Experimental Setup}
{\setlength{\parindent}{0cm}\textbf{Datasets.}}
For consistency with prior work \cite{pmlr-v162-xu22k, pmlr-v202-xu23f}, we evaluate our \sysname on the DeepCAD \cite{wu2021deepcad} dataset.
This dataset comprises 178,238 sketch-and-extrusion sequences, divided randomly into training, validation, and test sets in a ratio of 90\%-5\%-5\%. 
To ensure data quality, we follow SkexGen \cite{pmlr-v162-xu22k} to remove duplicate and invalid sequences.
Subsequently, we convert all resultant CAD sequences into texts, as mentioned in Sec. \ref{3.1}.

{\setlength{\parindent}{0cm}\textbf{Implementation Details.}}
We adopt the transformers \cite{wolf2020transformers} toolbox and select Llama-3-8B \cite{meta2024introducing} as the base LLM, which achieves superior performance among open-source LLMs.
For the 8B model, we use LoRA \cite{hu2022lora} to fine-tune only 0.042\%  of their parameters, approximately 3.4 million.
The LoRA rank and alpha are set to 8 and 32.
The model is trained on four A6000 GPUs.
we employ the AdamW optimizer \cite{loshchilov2018decoupled}, set the batch size to 32, use a cosine annealing learning rate of $5 \times 10^{-4}$, and train for 30 epochs. 
During the inference process, we set the sampling temperature $\tau$ and $\mathrm{Top\text{-}p}$ at 1.1 and 0.9, respectively.

{\setlength{\parindent}{0cm}\textbf{Metrics.}}
We adopt metrics consistent with
previous methods \cite{pmlr-v162-xu22k,pmlr-v202-xu23f}.
Generally, the metrics $\mathrm{Coverage \ (COV)}$, 
$\mathrm{Minimum \ Matching \ Distance \ (MMD)}$ and $\mathrm{Jensen\text{-}Shannon \ Divergence \ (JSD)}$ measure generation diversity and quality on generated CAD models in comparison to the test set.
See their detailed descriptions in \cite{pmlr-v202-xu23f}.
$\mathrm{Novel}$ indicates the percentage of generated CAD models not present in the training set, while $\mathrm{Unique}$ represents the percentage of generated CAD models that appear only once within the generated set.
$\mathrm{Prediction \ Validity \ (PV)}$ denotes the overall validity of predictions that can be rendered into 3D shapes rather than just 2D sketches or nothing.
$\mathrm{Realism}$ denotes the realistic rate of generated CAD models compared to the training data, as assessed by human evaluators.

\subsection{Performance Comparision with Existing Methods}\label{4.2}
{\setlength{\parindent}{0cm}\textbf{Baselines and Tasks.}}
We compare our \sysname with GPT-4o \cite{achiam2023gpt}, one of the most powerful closed-source LLMs, and
two state-of-the-art SEM-based baselines: SkexGen \cite{pmlr-v162-xu22k} and Hnc-cad \cite{pmlr-v202-xu23f}. 
Since SkexGen and Hnc-cad cannot simultaneously control CAD generation across all hierarchies as ours,
we choose the sketch-level and extrusion-level controllable generation tasks for comparison following the principles below.
First, they are common tasks that can be handled by each baseline. 
Second, there are official implementations of baselines for these tasks.
For GPT-4o, Hnc-cad and \sysname, given a CAD model from the test set, we randomly mask either a sketch or an extrusion and predict the corresponding masked field.
 Notably, despite our best efforts, SkexGen still has slightly different task settings compared to the above three methods.
Specifically, when there are multiple sketches (or extrusions) in a CAD model, SkexGen changes all the sketches (or extrusions) instead of the specified one.

{\setlength{\parindent}{0cm}\textbf{Quantitative Comparison Results.}}
\begin{table*}[!t]
  \centering
    \caption{Performance comparison on the DeepCAD test set.
    GPT-4o is enhanced with few-shot in-context learning. 
    Specifically, each prompt comprises five exemplars randomly chosen from the training set. 
These exemplars include instructions and answers, following the format shown in \fref{fig4}.
Moreover, we obtain the performance values of SkexGen and Hnc-cad based on their official codes.
    Best performances are in \textbf{bold}, and the second-bests are marked by *.
    }
  \setlength{\tabcolsep}{0.9pt}
  \renewcommand{\arraystretch}{1.31}
  \scalebox{0.8} {
    \begin{tabular}{c|cclccc|c|cclccc|c}
          \specialrule{\heavyrulewidth}{0pt}{0pt}
    \multicolumn{1}{c}{\multirow{2}{*}{Method}} & \multicolumn{7}{|c|}{Sketch-level}                      & \multicolumn{7}{c}{Extrusion-level} \\
    \cline{2-15} 
          & COV$\uparrow$ & MMD$\downarrow$ & JSD$\downarrow$& Novel$\uparrow$& Unique$\uparrow$ & PV$\uparrow$&Realism$\uparrow$ &  COV$\uparrow$ & MMD$\downarrow$& JSD$\downarrow$ & Novel$\uparrow$& Unique$\uparrow$& PV$\uparrow$& Realism$\uparrow$ \\
    \specialrule{\heavyrulewidth}{0pt}{0pt}
    GPT-4o & 58.2\% & 1.34 & 1.43 & 69.7\% & 72.8\% & 62.3\% & 23.2\%  & 53.3\% & 1.42 & 2.14 &  58.6\% & 65.3\% & 48.8\%& 19.7\% \\
    \hline
    SkexGen & 60.6\% & 1.27 & 1.51 & {90.7\%}* & \textbf{93.5\%} & 68.7\% & 34.8\%  & 63.6\% & 1.23 & 1.44 & \textbf{89.3\%} & 89.1\%* & 76.1\%& 35.2\% \\
    Hnc-cad & 62.4\%*  & 1.21* & 1.07*  &  87.6\% & 92.1\%  &  72.6\%* &  36.3\%*  & 65.6\%*  & 1.25*  & 1.38* &  86.2\% &  87.8\% &  79.7\%* &  38.0\% * \\
    Ours  & \textbf{65.6\%} & \textbf{1.19} & \textbf{0.82} & \textbf{92.1\%} & 92.6\%* & \textbf{93.4\%} & \textbf{39.6\%}  & \textbf{68.5\%} & \textbf{1.19} & \textbf{1.32} & 87.6\%* & \textbf{90.4\%} & \textbf{93.3\%}& \textbf{42.1\%}  \\
    \specialrule{\heavyrulewidth}{0pt}{0pt}
    \end{tabular}%
    } 

  \label{tab1}%
  \end{table*}%
\begin{figure*}[!t]
\centering
\includegraphics[width=\textwidth]{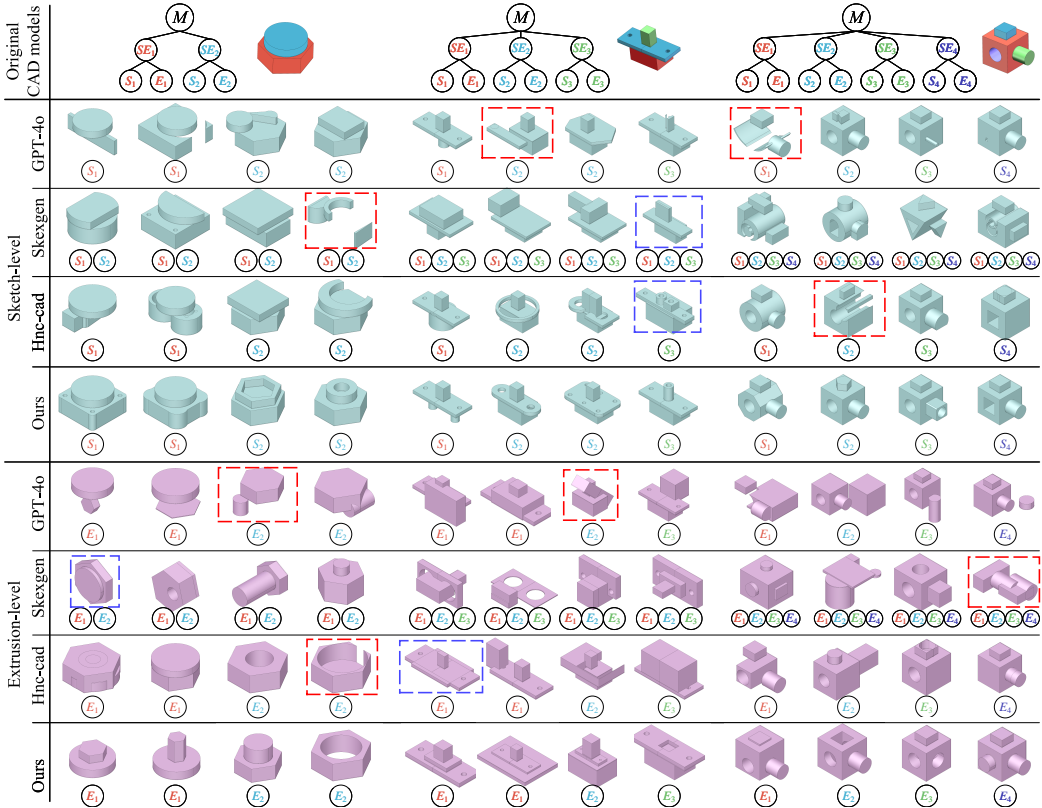}
\caption{Qualitative comparison results for four methods.
The first row displays three original CAD models, 
where the color of each sketch-extrusion aligns with that in the corresponding structural diagrams.
In the following rows, given a CAD model, we randomly select its four newly predicted models for each method. 
The marks below the predictions are the corresponding masked and modified sketches or extrusions.
The \textcolor{red}{red} boxes illustrate some of the most unrealistic examples. 
The \textcolor{b2}{blue} boxes indicate some of the most obvious cases, where multiple fields simultaneously change in the same CAD model, rather than just the expected masked field.
}
\label{fig5}
\end{figure*}
We randomly selected 1k CAD models from the test set. 
For every method, we generated 10 predictions for each model, resulting in a total of 10k CAD models. 
For the metrics $\mathrm{COV}$, $\mathrm{MMD}$, and $\mathrm{JSD}$ that require a subset of ground truths, we sampled 3k CAD models from the test set.
The average scores across three runs are presented in \tabref{tab1}. 
Without fine-tuning, GPT-4o performs poorly.
On the other hand, our \sysname outperforms the baselines on nearly all evaluation metrics, demonstrating significant superiority in generation quality and controllability.
Particularly, FlexCAD achieves the most notable improvement on $\mathrm{PV}$,
reaching up to 20.8\%-31.1\% and 13.6\%-44.5\% in terms of sketch-level and extrusion-level controllable generation.

{\setlength{\parindent}{0cm}\textbf{Qualitative Comparison Results.}}
To illustrate the performance intuitively, we randomly selected three CAD models from the test set. 
As shown in \fref{fig5}, the results clearly illustrate that \sysname greatly enhances the quality and controllability of CAD models compared to other competitors. 
Specifically, 
we tend to generate well-structured CAD models that closely resemble real-world examples, contrasting with unrealistic models like the red boxes shown in \fref{fig5}.
On the other hand, as shown in the blue boxes in \fref{fig5},
SkexGen cannot specify which sketch or extrusion to modify, 
and Hnc-cad cannot preserve the integrity of the unmasked elements, even when recovering from the same codes.
In contrast, we can mask any sketch or extrusion, ensuring that only the masked sketch or extrusion is modified while the remaining elements stay unchanged. 
This further confirms the effectiveness and superior controllability of our \sysname.

{\setlength{\parindent}{0cm}\textbf{Human Evaluation.}}
To evaluate $\mathrm{Realism}$, for each method, seven crowd workers were shown 950 pairs of images from the generated data and the training data, following \cite{pmlr-v202-xu23f}. 
They were asked to judge which of the two was more realistic.
As shown in \tabref{tab1}, for our \sysname, 39.6\% and 42.1\% of the generated models are more realistic.
These rates are the highest among all methods, further demonstrating the advantage of our \sysname.

\subsection{Enabling More Controllable Generation Tasks}
\begin{figure*}[!t]
\centering
\includegraphics[width=\textwidth]{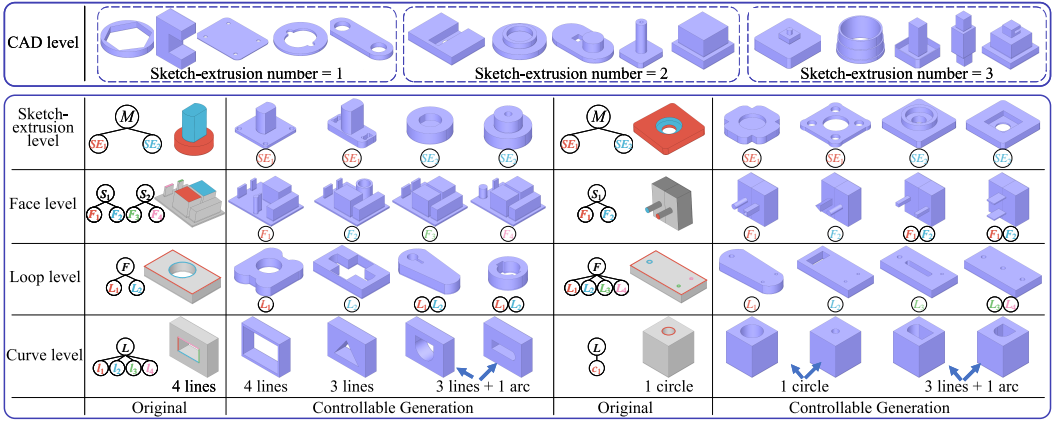}
\caption{Our \sysname achieves controllable generation across different hierarchies, as introduced in \ref{3.2}.
For the CAD level, we produce CAD models aligning with the required sketch-extrusion number.
For the sketch-extrusion, face and loop levels, 
the left side of each sub-figure shows an original CAD model along with its local structural diagram.
The color of each highlighted field matches that in the diagrams.
The right side shows the predictions with only the masked part being masked and edited.
And the masked part is marked below the predictions.
Similarly, for the curve level, below the predictions are user-defined curve type and number.
Best viewed in color.
}
\label{fig7}
\end{figure*}

In addition to the sketch and extrusion levels, our \sysname achieves controllable generation in other hierarchies, including the CAD, sketch-extrusion, face, loop, and curve levels.
We provide some examples in \fref{fig7}, where users can modify the specific fields of a CAD model according to their intent.
Additionally, we present detailed qualitative results across these hierarchies in \tabref{th} in the appendix.
All the results together illustrate the effectiveness of our \sysname.

\begin{figure*}[!t]
\centering
\includegraphics[width=\textwidth]{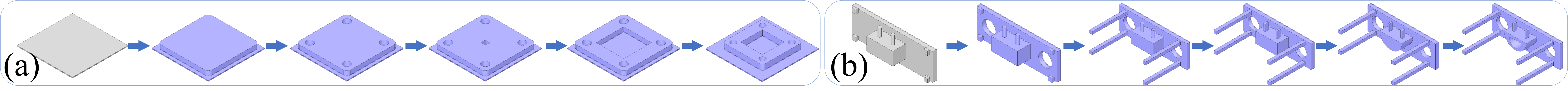}
\caption{
Two examples of iterative editing.
(a) Based on a simple CAD model, a new sketch-extrusion is generated by adding a $\mathrm{[sketch\text{-}extrusion \ mask]}$ at the end of the CAD text. Similarly, four peripheral internal loops are created. Then, a central quadrilateral loop is added and its geometry is altered. 
Finally, the new extrusion is adjusted to better match the original model.
(b) Based on a complex CAD model, modifications can be progressively applied at various sketch-extrusions with the loop-extrusion-face-curve-loop level controllable generation to ultimately meet user needs.
}
\label{fig8}
\end{figure*}

Furthermore, our \sysname can achieve iterative editing. 
For example, as shown in \fref{fig8}(a), 
starting from a simple CAD model, we iteratively edit it within a newly generated sketch-extrusion until the sketch-extrusion aligns closely with user requirements.
On the other hand, 
as shown in \fref{fig8}(b), given a complex CAD model, we continuously adjust diverse hierarchy elements within different sketch-extrusions until they progressively match user specifications.

\subsection{Ablation Studies}
We conduct several ablation studies evaluated on the sketch-level controllable generation as mentioned in Sec. \ref{4.2}, unless otherwise stated.

{\setlength{\parindent}{0cm}\textbf{Settings of LLMs.}}
\begin{table}[!t]
  \centering
  \caption{Ablation studies for fine-tuning LLMs with different settings.
  $\mathrm{Pre\text{-}trained}$ denotes the initial pre-trained weights.
  $\mathrm{Full}$ and $\mathrm{LoRA}$ indicate full and a few parameters are trainable, respectively.
  $\mathrm{Transformer\text{-}4M}$ 
  is a small transformer-based \cite{transformer} language model.
 Its total number of trainable parameters is comparable to that of our model with LoRA.
$\mathrm{Llama\text{-}3\text{-}8B\text{-}From\text{-}Scratch}$ and $\mathrm{Llama\text{-}3\text{-}8B\text{-}Full}$ denote training full parameters without and with the initial pre-trained weights, respectively.
$\mathrm{Llama\text{-}3\text{-}8B\text{-}Instruct}$ is an instruction-tuned model in an 8B size \cite{meta2024introducing}.
For $\mathrm{Llama\text{-}3\text{-}70B}$ \cite{meta2024introducing}, we fine-tune only 0.023\% of its parameters, around 16.3 million.
Best performances are in \textbf{bold} and the second-bests are marked by $*$.
  }
  \setlength{\tabcolsep}{4pt}
  \renewcommand{\arraystretch}{1.2}
  \scalebox{0.9} {
    \begin{tabular}{r|cccccc}
    \specialrule{\heavyrulewidth}{0pt}{0pt}
  Model  & COV$\uparrow$ & MMD$\downarrow$ & JSD$\downarrow$& Novel$\uparrow$& Unique$\uparrow$ &  PV$\uparrow$  \\
  \specialrule{\heavyrulewidth}{0pt}{0pt}
     Transformer-4M (w/o Pre-trained, Full)   &  59.4\%     &  1.37     &   1.02   & 85.8\%     &  86.9\%    & 80.2\% \\
       Llama-3-8B-From-Scratch (w/o Pre-trained, Full)   &  63.0\%    & 1.23     & 0.91     & 89.7\% &    90.2\%   &  89.5\%      \\
        Llama-3-8B-Full (w/ Pre-trained, Full) &       66.4\%*&   1.20   &  0.85    &  92.6\%*  &   91.0\%     &  91.7\%     \\
        Llama-3-8B-Instruct (w/ Pre-trained, LoRA)  &  65.3\%     &  1.22   &    0.89   &    91.4\%   &  92.1\%*     & 90.5\% \\
         Llama-3-8B (w/ Pre-trained, LoRA, ours) & 65.6\% & 1.19* & \textbf{0.82} & 92.1\% & \textbf{92.6\%} & 93.4\%*    \\
        Llama-3-70B (w/ Pre-trained, LoRA) &  \textbf{68.2\%}     &  \textbf{1.13}     &  0.84*      &  \textbf{93.0\%} & 91.8\%      & \textbf{94.6\%}     \\
          \specialrule{\heavyrulewidth}{0pt}{0pt}
    \end{tabular}%
    }
  \label{tm1}%
\end{table}%
As shown in \tabref{tm1}, without the pre-trained weights, both $\mathrm{Transformer\text{-}4M}$ and $\mathrm{Llama\text{-}3\text{-}8B\text{-}From\text{-}Scratch}$ achieve the lowest performances.
This indicates that the pre-trained weights in LLMs contain valuable knowledge, which contributes to the performance gains.
$\mathrm{Llama\text{-}3\text{-}8B\text{-}Full}$ achieves a performance similar to that of our model. However, it requires 80 hours to reach model convergence, compared to just 20 hours for ours. 
This highlights the effectiveness of the LoRA strategy \cite{hu2022lora}.
The performance of $\mathrm{Llama\text{-}3\text{-}8B\text{-}Instruct}$ is slightly lower than that of our model. 
Conversely, as the model scale increases significantly, $\mathrm{Llama\text{-}3\text{-}70B}$ achieves the best performance but is more time-consuming and costly (See Sec.~\ref{alf} in the appendix).

{\setlength{\parindent}{0cm}\textbf{Effectiveness of Hierarchy-Aware Masking.}}
\begin{table}[!t]
  \centering
  \caption{Effectiveness analysis of the hierarchy-aware masking strategy and unified training.
  $\mathrm{Random \ Masking}$ denotes randomly masking 15\%-50\% continuous tokens within each CAD text, instead of the hierarchy-aware field.
  $\mathrm{w/o \ Hierarchy\text{-}specific \ Tokens}$ 
  indicates that when masking, we utilize the generic token $\mathrm{[mask]}$, rather than employing hierarchy-specific mask tokens, such as $\mathrm{[face \ mask]}$, $\mathrm{[loop \ mask]}$ and \etc
  $\mathrm{w/o \ Unified \ Training}$ represents that we solely train a single task, \ie the sketch-level controllable generation.
  Best performances are in \textbf{bold}.
  }
  \setlength{\tabcolsep}{10pt}
  \renewcommand{\arraystretch}{1.2}
  \scalebox{0.9} {
    \begin{tabular}{r|cccccc}
    \specialrule{\heavyrulewidth}{0pt}{0pt}
  Model  & COV$\uparrow$ & MMD$\downarrow$ & JSD$\downarrow$& Novel$\uparrow$& Unique$\uparrow$ &  PV$\uparrow$  \\
  \specialrule{\heavyrulewidth}{0pt}{0pt}
  Random Masking  &   63.0\%    &  1.25   &   1.02    &  88.2\%      &  91.5\%    &  90.6\%  \\
     % Fixed Masking  & 64.2\%  & 1.20  &  0.91 & 91.8\% &  91.3\% &  91.2\%    \\
        w/o Hierarchy-specific Tokens  &  63.7\%     &  1.20   &    0.95   &    90.8\%   &  91.7\%     & 91.5\% \\
         w/o Unified Training & 64.3\% & \textbf{1.17} & 0.89 & 91.6\% & 90.9\% & 92.2\%    \\
        Ours &  \textbf{65.6\%} & 1.19 & \textbf{0.82} & \textbf{92.1\%} & \textbf{92.6\%} & \textbf{93.4\%}    \\
          \specialrule{\heavyrulewidth}{0pt}{0pt}
    \end{tabular}%
    }
  \label{tm2}%
\end{table}%
As shown in \tabref{tm2}, 
$\mathrm{Random \ Masking}$ exhibits the lowest performance, underscoring the effectiveness of masking the hierarchy-aware field. Furthermore, the absence of hierarchy-specific tokens leads to a performance decline.
In essence, these tokens potentially assist LLMs in more accurately determining the level they are generating, consequently enhancing overall controllable generation performance.

{\setlength{\parindent}{0cm}\textbf{Effectiveness of Unified Training.}}
As depicted in \tabref{tm2},
without unified training, there is a slight decrease in performance.
In other words, when simultaneously training multiple controllable generation tasks across various levels, the inter-task knowledge contributes to performance gains.

\section{Conclusion}
In this paper, we introduce a unified, versatile and user-friendly model, termed \sysname, which is particularly designed for controlling CAD generation across all hierarchies.
To the best of our knowledge, we are the first to utilize LLMs for controllable CAD generation.
Specifically, we convert each CAD model into a brief and structured text and propose hierarchy-aware masking for fine-tuning.
Our \sysname is simple yet highly effective. 
Thorough qualitative and quantitative assessments conducted on public benchmarks confirm its effectiveness across all hierarchies.

\section*{Ethics Statement}
The data used in this work is tailored for creating and modifying CAD models. Due to its specialized nature, the misuse risk is naturally minimized, ensuring that the developed methods primarily benefit design and engineering tasks. In this work, we have invited crowd workers to evaluate the quality of generated CAD models. We conducted this work in accordance with ethical guidelines to ensure that participants were treated fairly, respectfully, and safely throughout the process. We took steps to protect the privacy of crowd workers by not collecting personally identifiable information. The data annotated by the crowd workers was used only for research purpose related to improving CAD generating techniques.

\section*{Acknowledgement} 
In this work, Zhanwei Zhang, Wenxiao Wang and Deng Cai were supported in part by The National Nature Science Foundation of China (Grant No: 62303406, 62273302, 62036009, 61936006), in part by Zhiyuan Laboratory (NO. ZYL2024022b).

\bibliography{iclr2025_conference}
\bibliographystyle{iclr2025_conference}
\newpage
\appendix
{\huge{Appendix}}

Considering the space limitation of the main paper, we provided more results and discussion in this appendix, which is organized as follows: 
\begin{itemize}
\item Section \ref{a1}: Additional Implementation Details and Analysis.
\begin{itemize}
\item Sec.~\ref{atc}: Meaning of Numbers in The CAD Text.
% \item Sec.~\ref{atd}: Additional Training Details.
\item Sec.~\ref{ahe}: Detailed Results for Human Evaluation.
\item Sec.~\ref{aqr}: Quantitative results for Other Hierarchies.
\item Sec.~\ref{asa}: Sensitivity Analysis of Key Hyper-parameters in Sampling.
\item Sec.~\ref{aug}: Unconditional Generation Task.
\item Sec.~\ref{alf}: Limitations and Future Work.
\end{itemize}
\end{itemize}

\begin{itemize}
\item Section B: Additional Qualitative Results.
\begin{itemize}
\item Fig.~\ref{ac}: CAD Level Controllable Generation.
\item Fig.~\ref{ase}: Sketch-extrusion Level Controllable Generation.
\item Fig.~\ref{ae}: Extrusion level Controllable Generation.
\item Fig.~\ref{as}: Sketch Level Controllable Generation.
\item Fig.~\ref{af}: Face Level Controllable Generation.
\item Fig.~\ref{al}: Loop Level Controllable Generation.
\item Fig.~\ref{ali}: Curve Level Controllable Generation.
\item Fig.~\ref{aun}: Unconditional Generation.
\end{itemize}
\end{itemize}

\section{Additional Implementation Details and Analysis}\label{a1}
\subsection{Meaning of Numbers in The CAD Text.}\label{atc}
\begin{table}[h]
  \centering
  \caption{Effectiveness analysis of circle representation. To denote a circle,
  $\mathrm{Center\&Radius}$ utilizes the center coordinates along with the radius, while
  $\mathrm{Diameter}$ uses two uniformly distributed points on the circumference that collectively define the diameter.
  Best performances are in \textbf{bold}.
  }
  \setlength{\tabcolsep}{13.5pt}
  \renewcommand{\arraystretch}{1.2}
  \scalebox{0.9} {
    \begin{tabular}{r|cccccc}
    \specialrule{\heavyrulewidth}{0pt}{0pt}
  Model  & COV$\uparrow$ & MMD$\downarrow$ & JSD$\downarrow$& Novel$\uparrow$& Unique$\uparrow$ &  PV$\uparrow$  \\
  \specialrule{\heavyrulewidth}{0pt}{0pt}
        Center\&Radius  &  63.2\%     &  1.21   &    0.87   &    90.2\%   &  90.9\%     & 89.7\% \\
         Diameter & \textbf{66.4\%} & {1.20} & 0.87 & 90.7\% & 91.5\% & 90.3\%    \\
        Four points (Ours) &  {65.6\%} & \textbf{1.19} & \textbf{0.82} & \textbf{92.1\%} & \textbf{92.6\%} & \textbf{93.4\%}    \\
          \specialrule{\heavyrulewidth}{0pt}{0pt}
    \end{tabular}%
    }
  \label{acr}%
\end{table}%
In this part, we explain the meaning of numbers in the CAD text, which builds upon SkexGen \cite{pmlr-v162-xu22k}.
In sketches, two (three) points can actually form a line (arc).
In our work, a line, an arc, and a circle are represented by one, two, and four points, respectively.
Each point is denoted as its $x$ and $y$ coordinates.
Here, the second (third) point of a line (arc) is determined by the first point of the subsequent curve (or the first curve when a loop is closed).
Four points of a circle are uniformly distributed along the circumference.
We implement two variants of circle representation.
As shown in Table~\ref{acr}, when evaluated on the sketch-level controllable generation task, our \sysname displays robustness across different circle representations, with $\mathrm{Four\ points}$ showing a slight edge.

Each extrusion operation is represented by 18 parameters:
BVVTTTRRRRRRRRRSOO. \\
- B represents one of the three Boolean operations: add, cut or intersect. It occupies 1 parameter. \\
- V indicates the displacements of the top and the bottom planes from the reference plane in which a sketch is extruded to form a solid. It occupies 2 parameters. \\
- T represents the 3D translation applied to the extruded solid. It occupies 3 parameters. \\
- R represents the 3D rotation of the extrusion direction. It occupies 9 parameters. \\
- S represents the uniform scaling factor. It occupies 1 parameter. \\
- O represents the center of scaling as a 2D coordinate. It occupies 2 parameters.

\subsection{Detailed Results for Human Evaluation}\label{ahe}
\begin{figure*}[!t]
\centering
\includegraphics[width=\textwidth]{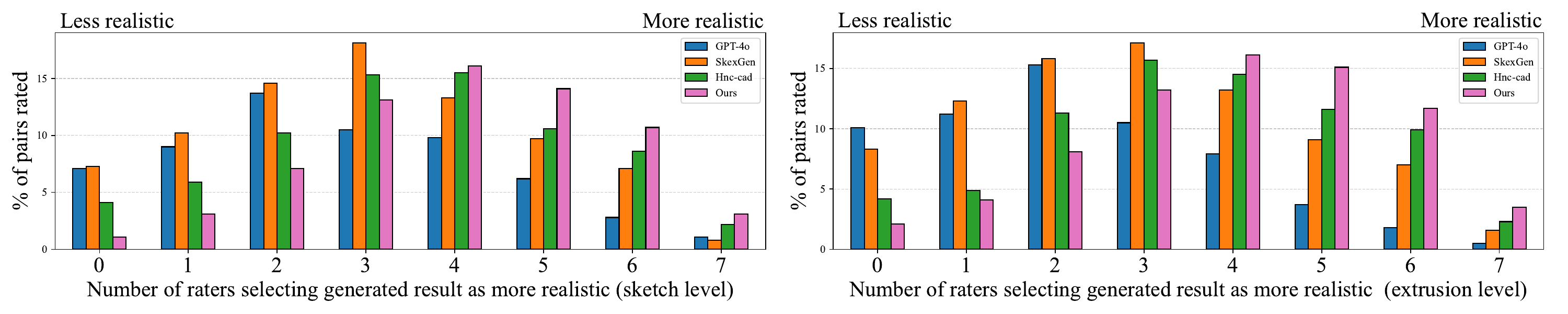}
\caption{Distribution of realism scores from seven human evaluators.
These scores are derived by comparing the generated CAD models produced by the four methods with the training samples.
}
\label{me}
\end{figure*} 
We present the detailed distribution of the $\mathrm{Realism}$ scores as mentioned in Table \ref{tab1}.
As shown in the \fref{me}, 
the distributions for GPT-4o, SkexGen, and Hnc-cad are skewed towards the `less realistic' end.
Conversely, our \sysname demonstrates a primarily symmetric distribution, suggesting that the crowd workers struggle to differentiate between the generated models and the training set.

\subsection{Quantitative results for Other Hierarchies}\label{aqr}
% \vspace{-0.5cm}
\begin{table}[h]
  \centering
  \caption{Quantitative results for controllable generation across other hierarchies.
The detailed evaluation setting is the same as that of 
Table \ref{tab1}.
  }
  \setlength{\tabcolsep}{12pt}
  \renewcommand{\arraystretch}{1.2}
  \scalebox{0.9} {
    \begin{tabular}{r|cccccc}
    \specialrule{\heavyrulewidth}{0pt}{0pt}
  Hierarchy  & COV$\uparrow$ & MMD$\downarrow$ & JSD$\downarrow$& Novel$\uparrow$& Unique$\uparrow$ &  PV$\uparrow$  \\
  \specialrule{\heavyrulewidth}{0pt}{0pt}
      CAD level    &  67.2\%     &   1.14    &    0.77   &    92.6\%   &  93.2\%     & 91.8\% \\
       Sketch-extrusion level   &   65.3\%    &    1.21   &  0.80     &    90.3\%   &   89.7\%    & 90.5\% \\
       Face level   &  62.9\%     &   1.18    &    0.84    & 91.1\%&    90.9\%   &  93.2\%      \\
         Loop level &   63.4\%    &  1.15     & 0.81       &  88.3\%&   85.7\%    &   90.5\%    \\
         Curve level &  59.1\%     &  1.20     &  0.79      &  89.7\% & 91.5\%      & 90.2\%      \\
          \specialrule{\heavyrulewidth}{0pt}{0pt}
    \end{tabular}%
    }
  \label{th}%
\end{table}%
In this section, we report the quantitative results across other hierarchies, including CAD, sketch-extrusion, face, loop and curve levels.
By combining the data from Table \ref{tab1} and Table \ref{th}, we observe that there is not a significant difference in performance across all hierarchies.
These results together illustrate the effectiveness of our FlexCAD across all hierarchies.

\subsection{Sensitivity Analysis of Key Hyper-parameters in Sampling}\label{asa}
\begin{table}[h]
  \centering
  \caption{Effectiveness analysis of key hyper-parameters, including the sampling temperature $\tau$ and $\mathrm{Top\text{-}p}$.
  Best performances are in \textbf{bold} and the second-bests are marked by *.
  }
  \setlength{\tabcolsep}{14.2pt}
  \renewcommand{\arraystretch}{1.2}
  \scalebox{0.9} {
    \begin{tabular}{c|cccccc}
    \specialrule{\heavyrulewidth}{0pt}{0pt}
  Model  & COV$\uparrow$ & MMD$\downarrow$ & JSD$\downarrow$& Novel$\uparrow$& Unique$\uparrow$ &  PV$\uparrow$  \\
  \specialrule{\heavyrulewidth}{0pt}{0pt}
       $\tau = 0.9$ &  {62.7\%} & 1.20 & {0.84} & {92.0\%}* & {90.3\%} & \textbf{96.7\%}    \\
       $\tau = 1.1$ &  {65.6\%}* & 1.19* & {0.82}* & \textbf{92.1\%} & {92.6\%}* & {93.4\%}*    \\
       $\tau = 1.3$ &  \textbf{65.8\%} & \textbf{1.12} & \textbf{0.78} & {91.9\%} & \textbf{94.5\%} & {86.8\%}    \\
          \specialrule{\heavyrulewidth}{0pt}{0pt}
    $\mathrm{Top\text{-}p} = 0.8$ &  {62.9\%} & 1.23 & {0.90} & {91.7\%} & {87.8\%} & \textbf{95.2\%}    \\
   $\mathrm{Top\text{-}p} = 0.9$ &  {65.6\%}* & 1.19* & {0.82}* & {92.1\%}* & {92.6\%}* & {93.4\%}*    \\
   $\mathrm{Top\text{-}p} = 1.0$ &  \textbf{68.3\%} & \textbf{1.13} & \textbf{0.77} & \textbf{93.3\%} & \textbf{95.7\%} & {89.0\%}    \\
     \specialrule{\heavyrulewidth}{0pt}{0pt}
    \end{tabular}%
    }
  \label{hy}%
\end{table}%
In this part, we perform a sensitivity analysis on key hyperparameters in inference, including the sampling temperature $\tau$ and $\mathrm{Top\text{-}p}$.
As shown in Table \ref{hy}, 
as either $\tau$ or $\mathrm{Top\text{-}p}$ increases, the performance of the first five metrics exhibits improvement, whereas the performance of the last deteriorates. 
Essentially, higher values of $\tau$ or $\mathrm{Top\text{-}p}$ lead to predictions that are more random and varied, while the overall prediction validity $\mathrm{PV}$ declines.
In our experiments, we made a trade-off by selecting the values of $\tau$ and $\mathrm{Top\text{-}p}$ to guarantee that the $\mathrm{PV}$ value remains above 90\%.

\subsection{Unconditional Generation Task}\label{aug}

\begin{table}[h]
  \centering
  \caption{Performance comparison for the unconditional generation task. 
  Each method generates 10k CAD models, which are then compared with a randomly selected subset of 2.5k ground truth models from the test set.
  For the baselines, other than the metric $\mathrm{PV}$, we derive the values of other metrics from the original paper \protect\cite{pmlr-v202-xu23f}, while the values of $\mathrm{PV}$ are obtained based on their official codes.
  Best performances are in \textbf{bold}, and the second-bests are marked by *.
  }
  \setlength{\tabcolsep}{11.5pt}
  \renewcommand{\arraystretch}{1.2}
  \scalebox{0.9} {
    \begin{tabular}{c|cccccc|c}
    \specialrule{\heavyrulewidth}{0pt}{0pt}
  Model  & COV$\uparrow$ & MMD$\downarrow$ & JSD$\downarrow$& Novel$\uparrow$& Unique$\uparrow$ &  PV$\uparrow$ & Realism  \\
  \specialrule{\heavyrulewidth}{0pt}{0pt}
       SkexGen &  {84.7\%} & 1.02 & {0.90}* & {99.1\%}* & \textbf{99.8\%} & {74.2\%}   & 46.9\%  \\
       Hnc-cad &  {87.7\%}* & 0.96* & \textbf{0.68} & {93.9\%} & {99.7\%}* & {77.4\%}*  & 49.2\%  \\
       Ours &  \textbf{89.2\%} & \textbf{0.91} & {1.53} & \textbf{99.3\%} & {96.9\%} & \textbf{90.5\%}  & 51.5\%  \\
          \specialrule{\heavyrulewidth}{0pt}{0pt}
    \end{tabular}%
    }
  \label{uncon}%
\end{table}%
Our \sysname can easily achieve unconditional CAD generation by simply adding a prompt template during training. 
Specifically, given a CAD text, the instruction in the prompt template can be as concise as `\emph{Below is a description of a CAD sequence:}', while the corresponding answer is the whole CAD text.
The quantitative and qualitative results shown in Table \ref{uncon} and Figure \ref{aun} verify the effectiveness of our \sysname in unconditional CAD generation.
Notably, as shown in Table \ref{uncon}, our $\mathrm{JSD}$ exhibits the poorest performance. 
However, upon adjusting the sampling temperature $\tau$ or $\mathrm{Top\text{-}p}$ to maintain the $\mathrm{PV}$ value at around 80\%, the $\mathrm{JSD}$ value enhances significantly to 0.78.

\subsection{Limitations and Future Work}\label{alf}
{\setlength{\parindent}{0cm}\textbf{Inference time.}}
We measure the inference time on one A6000 GPU (with a batch size of 1) for the extrusion-level generation, averaging over 1,000 runs. 
The inference time of our \sysname (based on Llama-3-8B) is slightly higher than that of SkexGen \cite{pmlr-v162-xu22k} and Hnc-cad \cite{pmlr-v162-xu22k}, at 0.56 seconds compared to 0.15 seconds and 0.38 seconds, respectively. 
Notably, we trained and tested Llama-3-70B using four A100 GPUs (with a batch size of 1 per GPU), yet the average inference time is still close to 3 seconds.
Although LLMs demonstrate promising performance, they generally lack efficiency. 
while the task of controllable CAD generation is not particularly demanding in terms of real-time inference requirements, the slight increase in inference time of our \sysname (based on Llama-3-8B) is acceptable given the promising performance.

\begin{figure*}[h]
\centering
\includegraphics[width=0.9\textwidth]{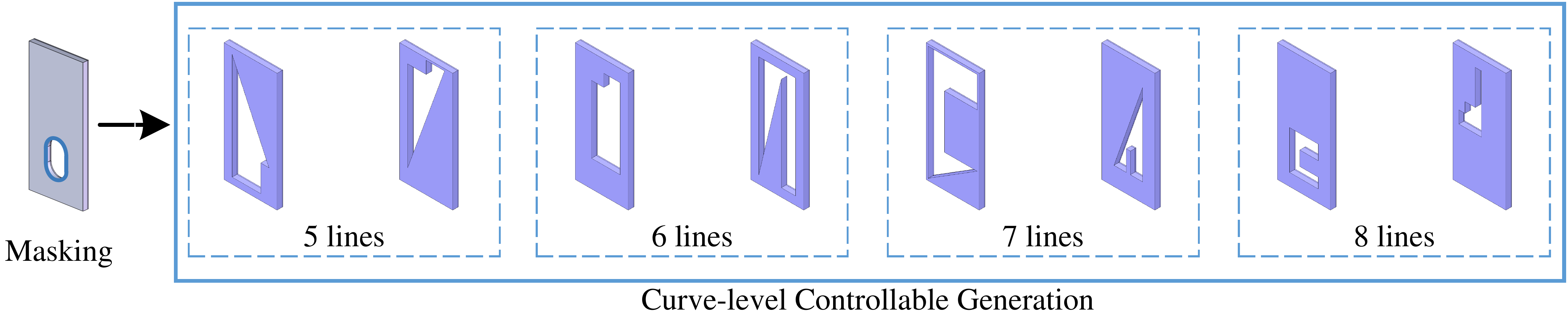}
\caption{Failure cases. 
The generated CAD models display irregular polygonal shapes.
}
\label{afa}
\end{figure*}
{\setlength{\parindent}{0cm}\textbf{Failure cases.}}
Despite the significant progress, our \sysname sometimes generates hallucinations. 
For example, when we implement curve-level controllable generation by replacing the original loops with 5, 6, 7, or 8 lines, our \sysname tends to generate plausible CAD texts. 
The CAD models produced from these texts typically exhibit irregular polygonal shapes, as illustrated in \fref{afa}.
Interestingly, when the modifications are limited to 4 lines, the generated predictions tend to be more structurally regular, as illustrated in \fref{ali}.
We select loops composed of 4, 5, 6, 7, or 8 lines in the training set for analysis.
The ratio is approximately 77.9\%-6.2\%-9.5\%-1.7\%-4.7\%. 
Consequently, the fundamental reason behind these hallucinations may stem from data imbalance, which could potentially be mitigated by collecting additional data.

{\setlength{\parindent}{0cm}\textbf{Other limitations and future work.}}
In this paper, we fine-tune LLMs to achieve controllable CAD generation.
However, several unexplored domains warrant investigation.
One such domain involves determining how LLMs can be effectively utilized for text-based generation tasks, 
such as providing a descriptive text like `\emph{Create a desk.}' to generate a corresponding CAD model.
Since LLMs possess robust capabilities in controllable CAD generation and text comprehension, in future work,
we aim to build a more advanced multi-modal LLM tailored for the text-based CAD generation task.

\begin{figure*}[h]
\centering
\includegraphics[width=\textwidth]{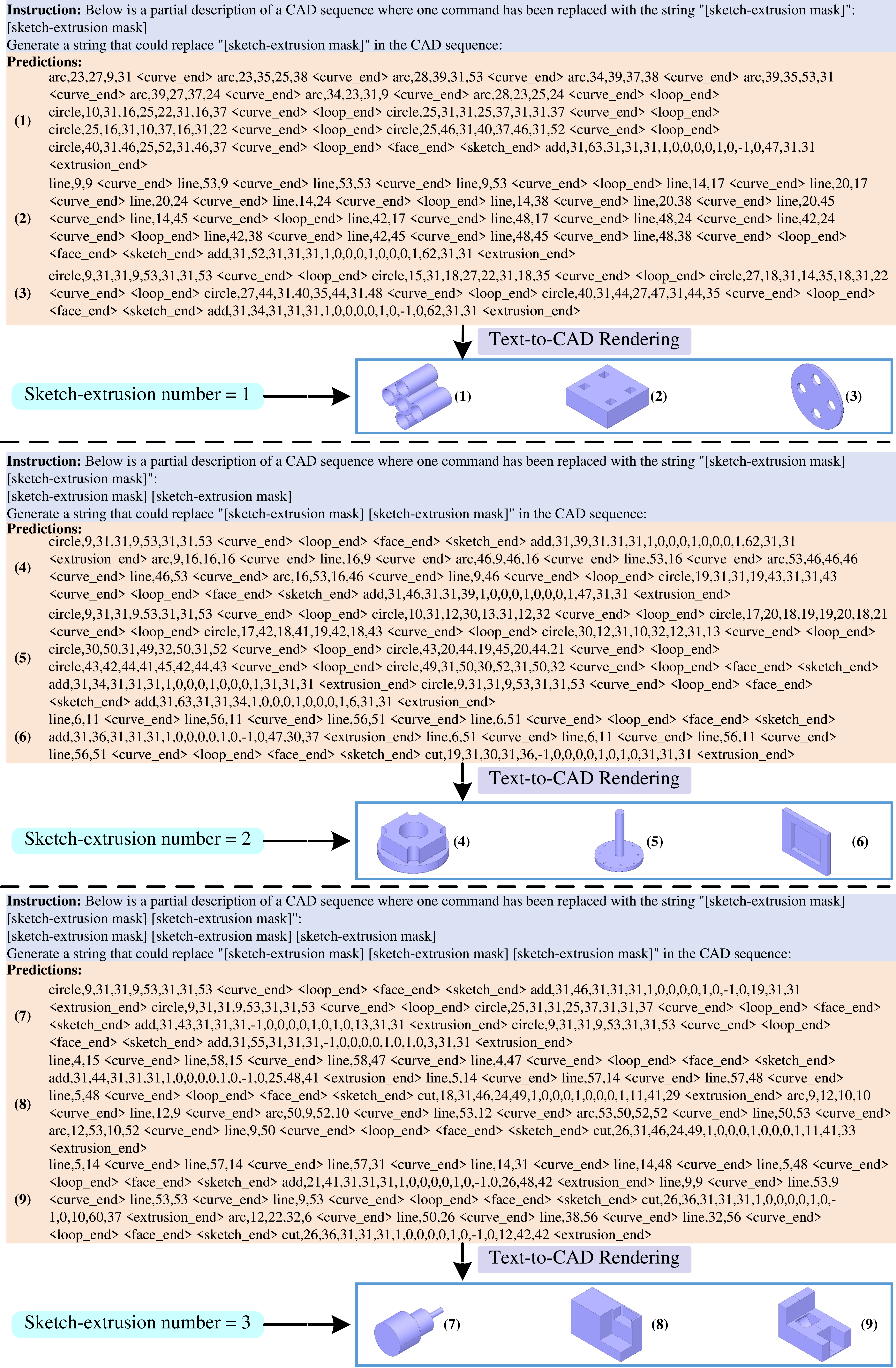}
\caption{CAD level generation.
Here, users can produce CAD models that match the required sketch-extrusion number. 
The marks of predictions align with those of the images (the same below).
}
\label{ac}
\end{figure*}

\begin{figure*}[h]
\centering
\includegraphics[width=\textwidth]{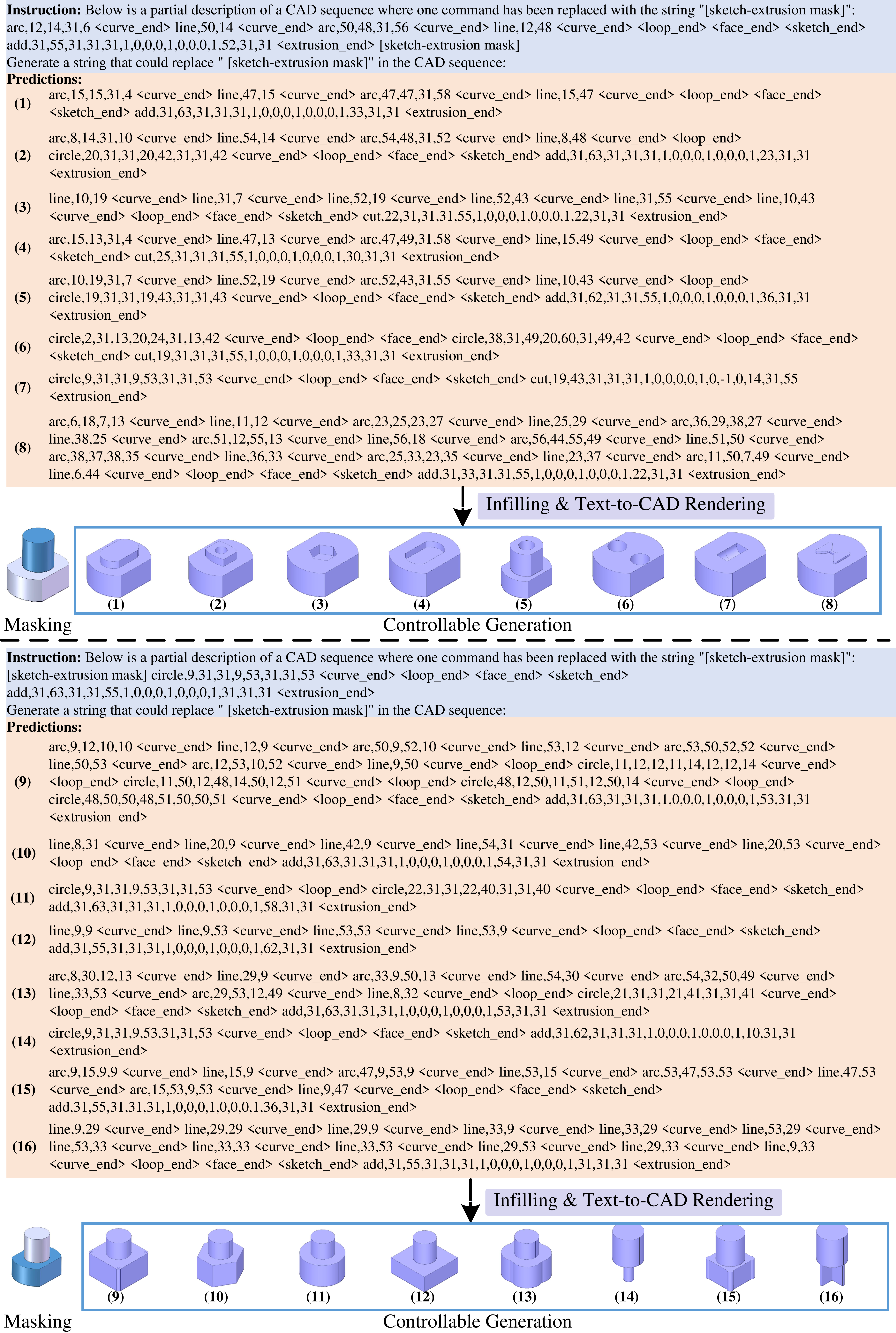}
\caption{Sketch-extrusion level generation.
Given a CAD model, users can mask any of its sketch-extrusions for modifications.
The masked sketch-extrusions are highlighted in \textcolor{cf4}{blue}.
}
\label{ase}
\end{figure*}

\begin{figure*}[h]
\centering
\includegraphics[width=\textwidth]{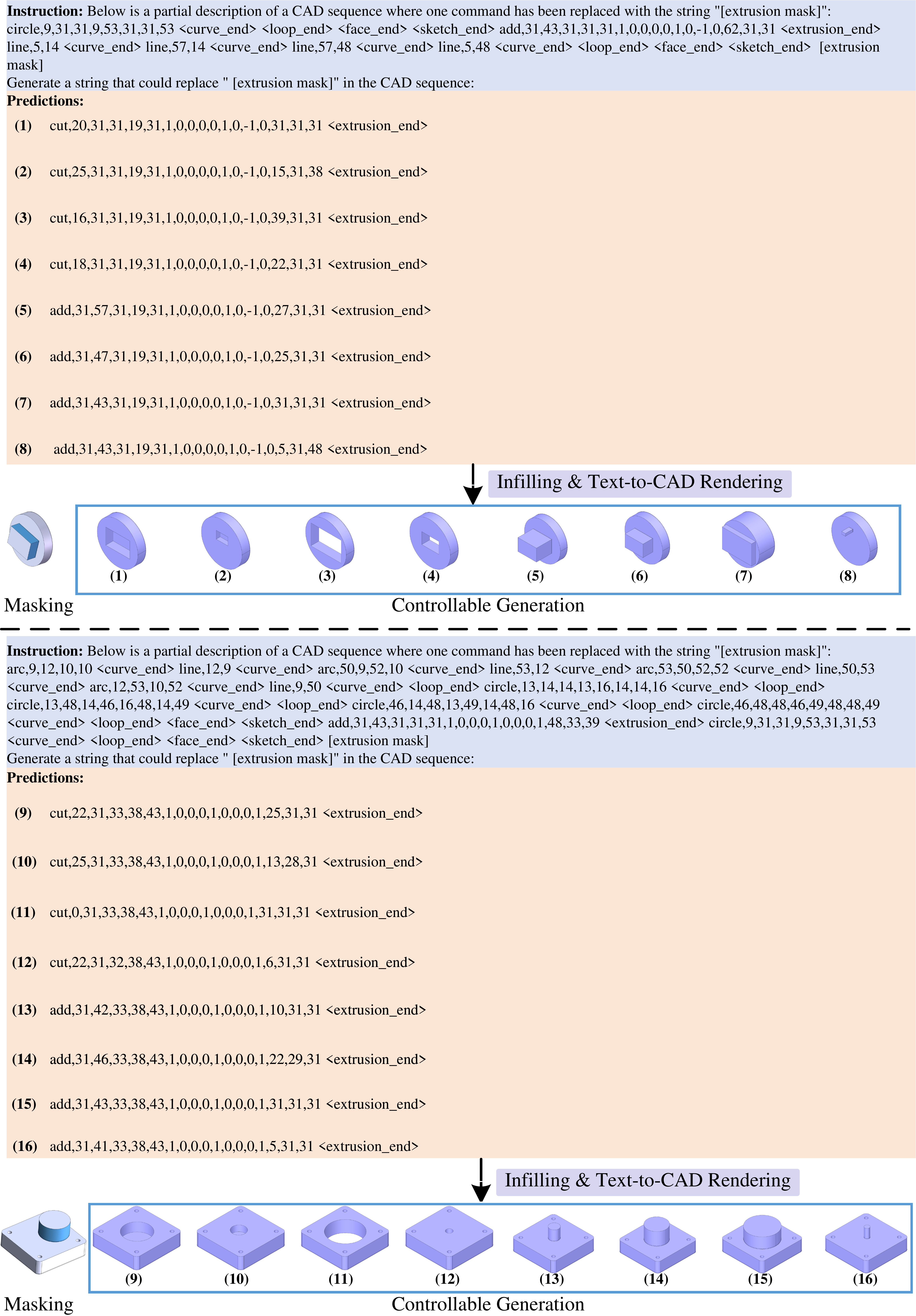}
\caption{Extrusion level generation.
Given a CAD model, users can mask any of its extrusions for modifications.
The masked extrusions are highlighted in \textcolor{cf4}{blue}.
}
\label{ae}
\end{figure*}

\begin{figure*}[h]
\centering
\includegraphics[width=\textwidth]{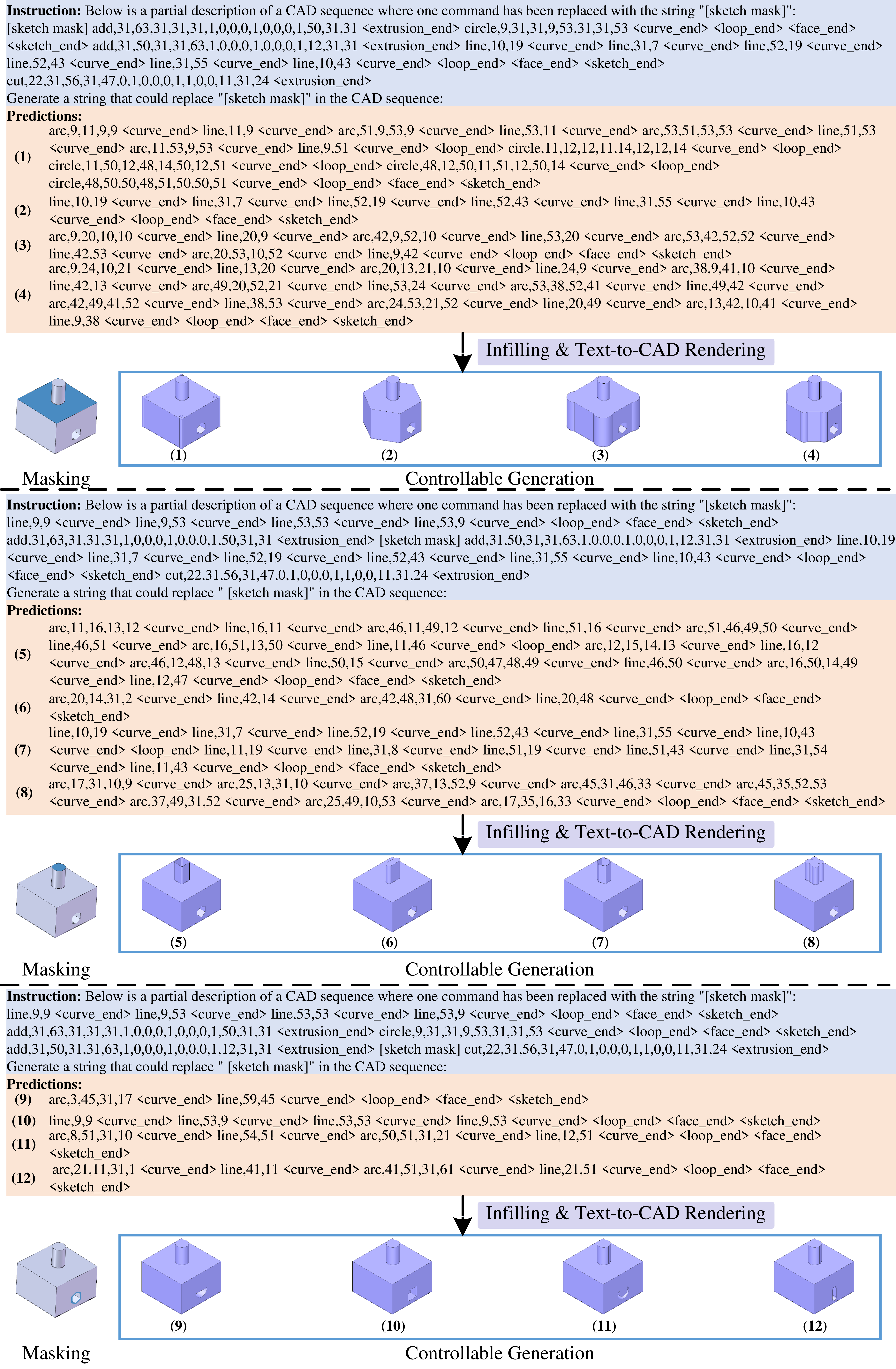}
\caption{Sketch level generation.
Given a CAD model, users can mask any of its sketches for modifications.
The masked sketches are highlighted in \textcolor{cf4}{blue}.
}
\label{as}
\end{figure*}

\begin{figure*}[h]
\centering
\includegraphics[width=\textwidth]{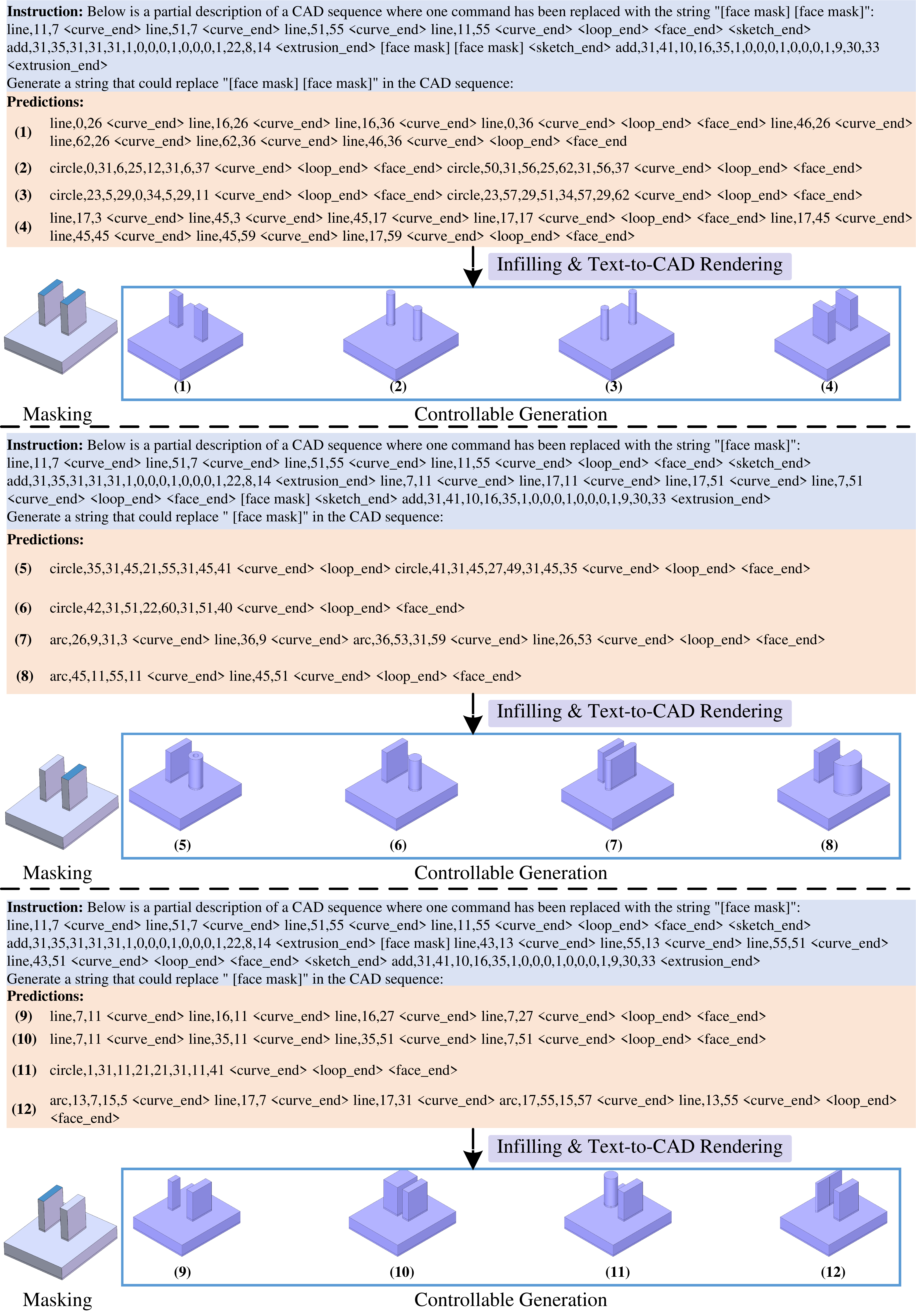}
\caption{Face level generation.
Given a sketch, users can mask one or multiple faces within it for modifications.
The masked faces are highlighted in \textcolor{cf4}{blue}.
}
\label{af}
\end{figure*}

\begin{figure*}[h]
\centering
\includegraphics[width=\textwidth]{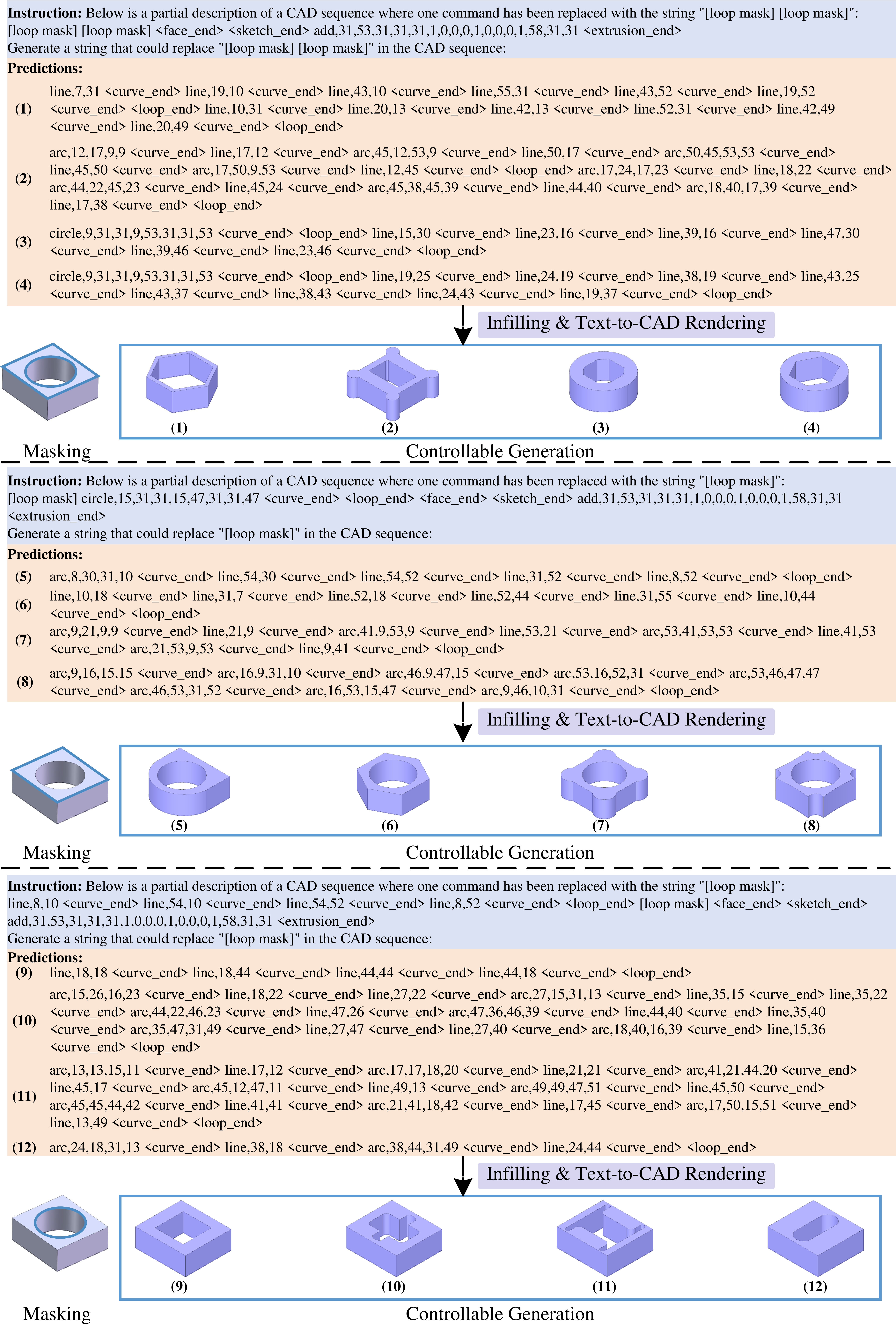}
\caption{Loop level generation.
Given a face, users can mask one or multiple loops within it for modifications.
The masked loops are highlighted in \textcolor{cf4}{blue}.
}
\label{al}
\end{figure*}

\begin{figure*}[h]
\centering
\includegraphics[width=\textwidth]{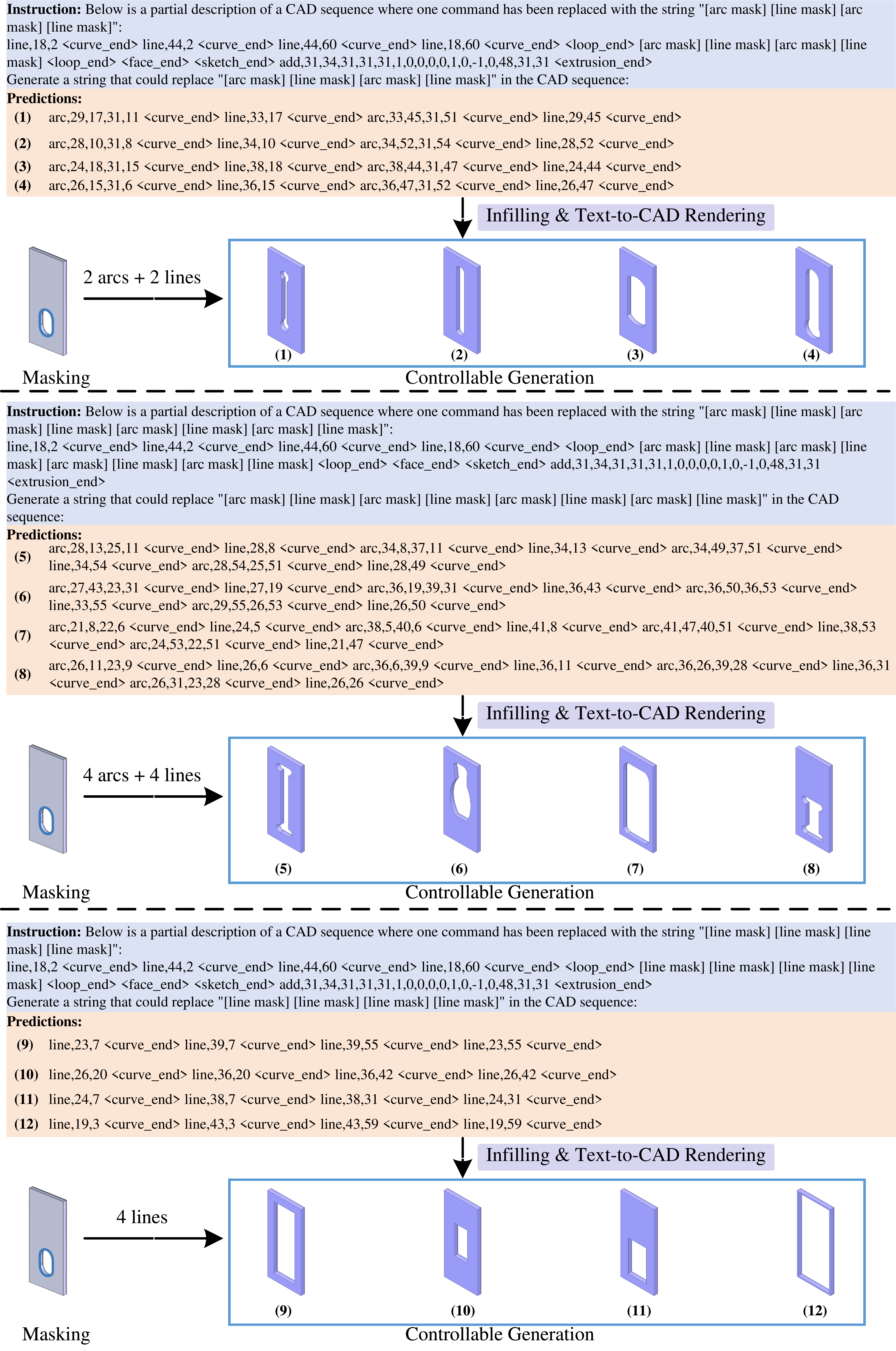}
\caption{Curve level generation. 
Given a loop, users can mask all curves within it and specify the curve type and number of newly generated loops.
The masked curves are highlighted in \textcolor{cf4}{blue}.
}
\label{ali}
\end{figure*}

\begin{figure*}[h]
\centering
\includegraphics[width=\textwidth]{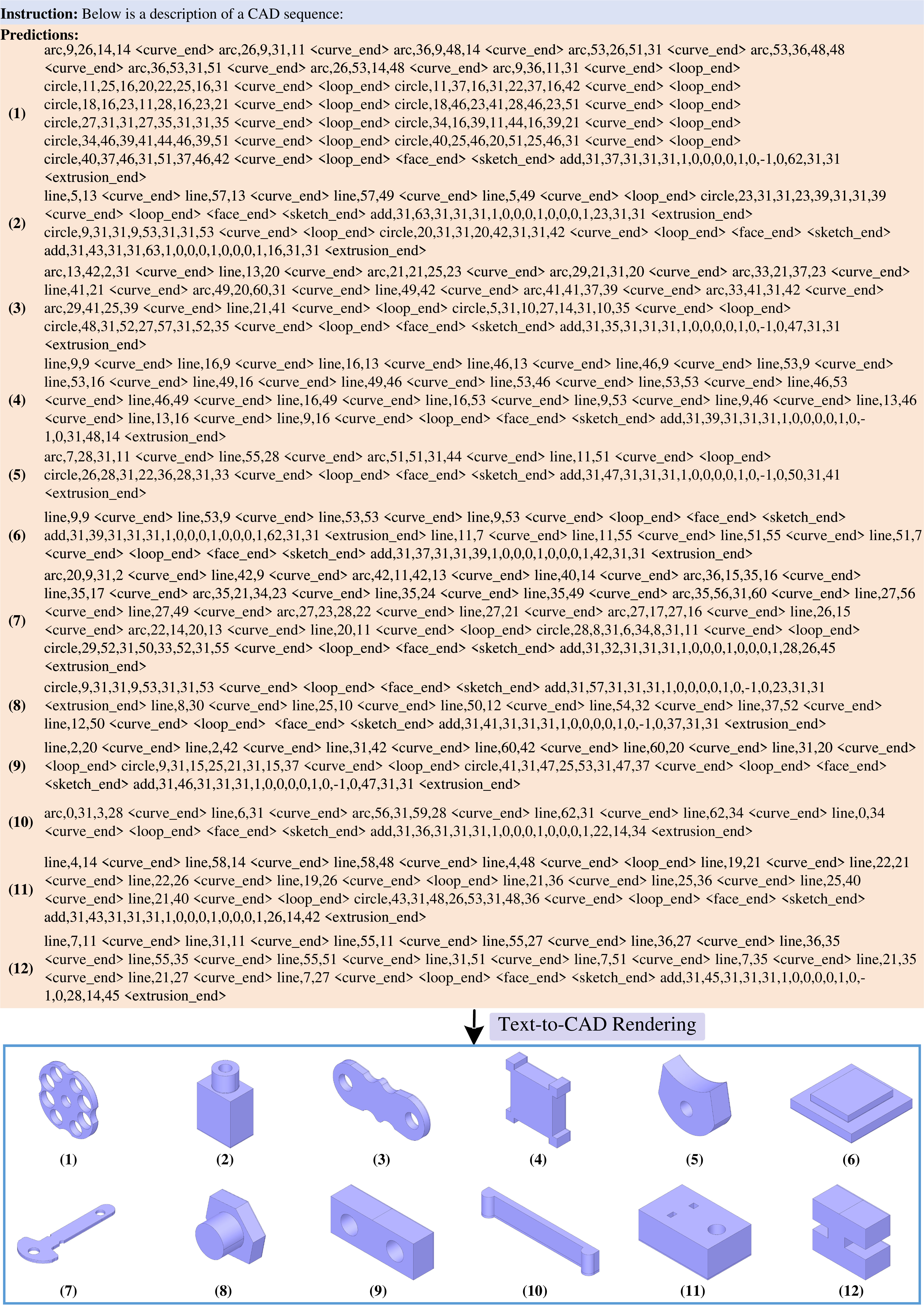}
\caption{Unconditional CAD generation.
Here, users can generate CAD models without any conditional limitations.
}
\label{aun}
\end{figure*}

\end{document}